\documentclass[10pt, conference, letterpaper]{IEEEtran}
\IEEEoverridecommandlockouts
\usepackage{cite}
\usepackage{color}
\usepackage{amsmath,amssymb,amsfonts}
\usepackage{algorithmic}
\usepackage{graphicx}
\usepackage{textcomp}
\usepackage{tcolorbox}
\tcbuselibrary{most}
\usepackage{CJKutf8}
\def\BibTeX{{\rm B\kern-.05em{\sc i\kern-.025em b}\kern-.08em
    T\kern-.1667em\lower.7ex\hbox{E}\kern-.125emX}}

\usepackage[ruled]{algorithm2e}

\usepackage{url}

\def\BibTeX{{\rm B\kern-.05em{\sc i\kern-.025em b}\kern-.08em T\kern-.1667em\lower.7ex\hbox{E}\kern-.125emX}}

\newcommand {\mymarginpar}[1]{\marginpar{#1}}
\renewcommand {\marginpar}[1]{}

\def\_{\rule{.3em}{.15ex}}      

\newcommand{\ls}[1]
   {\dimen0=\fontdimen6\the\font
    \lineskip=#1\dimen0
    \advance\lineskip.5\fontdimen5\the\font
    \advance\lineskip-\dimen0
    \lineskiplimit=.9\lineskip
    \baselineskip=\lineskip
    \advance\baselineskip\dimen0
    \normallineskip\lineskip
    \normallineskiplimit\lineskiplimit
    \normalbaselineskip\baselineskip
    \ignorespaces
   }

\newcommand {\bearn}{\begin{eqnarray*}}
\newcommand {\eearn}{\end{eqnarray*}}
\newcommand {\barr}{\begin{array}}
\newcommand {\earr}{\end{array}}

\renewcommand {\L}{{\cal L}}

\newcommand {\N}{{\cal N}}

\newtheorem{definition}{Definition}
\newtheorem{property}[definition]{Property}
\newtheorem{proposition}[definition]{Proposition}
\newtheorem{lemma}[definition]{Lemma}
\newtheorem{theorem}[definition]{Theorem}
\newtheorem{corollary}[definition]{Corollary}
\newtheorem{example}{Example}
\newtheorem{remark}[definition]{Remark}

\newcommand {\benum} {\begin{enumerate}}
\newcommand {\eenum} {\end{enumerate}}

\newcommand {\bdesc} {\begin{description}}
\newcommand {\edesc} {\end{description}}

\newcommand {\bfig}[2] {\begin{figure}
  \centering
  \includegraphics[width=#2]{#1}}
\newcommand {\brotatefig}[2] {\begin{figure}[htbp]
                        \centerline {
                         \epsfig{figure={#1},clip=,angle=-90,width={#2}}}}
\newcommand {\bfigfirst}[2] {\begin{figure}[h]
                        \centerline {
                        \setlength{\epsfxsize}{#2}
                        \epsffile{#1}}}
\newcommand {\efig}[2]{ \caption{#2}
                        \label{fig:#1}
                        \end{figure}
                        \mymarginpar{fig:#1}}
\newcommand {\erotatefig}[2]{ \caption{#2}
                        \label{fig:#1}
                        \end{figure}
                        \mymarginpar{fig:#1}}
\newcommand {\rfig}[1]{Figure \ref{fig:#1}}

\newcommand {\btab}[1]{
                       \begin{table}
                       \centering
                       \begin{tabular}{#1}}
\newcommand {\etab}[3] {
                       \end{tabular}
                       \caption[#3]{#2}
                       \label{tab:#1}
                       \end{table}
                       \mymarginpar{tab:#1}
                       \vspace{.1in}}

\newcommand {\btabular}[1]{\begin{center}
                       \begin{tabular}{#1}}
\newcommand {\etabular}{\end{tabular}
                       \end{center}}

\newcommand {\bdefin}[1]{\begin{definition}
                      \mymarginpar{def:#1}
                      \label{def:#1} }
\newcommand {\edefin}       {\end{definition}}

\newcommand {\bpro}[1]{\begin{property}
                      \mymarginpar{pro:#1}
                      \label{pro:#1} }
\newcommand {\epro}   {\end{property}}

\newcommand {\bprop}[1]{\begin{proposition}
                      \mymarginpar{prop:#1}
                      \label{prop:#1} }
\newcommand {\eprop}       {\end{proposition}}

\newcommand {\blem}[1]{\begin{lemma}
                      \mymarginpar{lem:#1}
                      \label{lem:#1} }
\newcommand {\elem}   {\end{lemma}}

\newcommand {\bthe}[1]{\begin{theorem}
                      \mymarginpar{the:#1}
                      \label{the:#1} }
\newcommand {\ethe}   {\end{theorem}}

\newcommand {\bcor}[1]{\begin{corollary}
                      \mymarginpar{cor:#1}
                      \label{cor:#1} }
\newcommand {\ecor}   {\end{corollary}}

\newcommand {\bax}[1]{\begin{axiom}
                      \mymarginpar{ax:#1}
                      \label{ax:#1} }
\newcommand {\eax}       {\vspace{-.1in} \end{axiom}}

\newcommand {\bex}[2]{\vspace{.1in}
                      \begin{example}
                      \mymarginpar{ex:#1}
                       {\bf #2}
                      \label{ex:#1} }
\newcommand {\eex}       {\end{example} \vspace{.3cm} }

\newcommand {\brem}[1]{\begin{remark}
                      \mymarginpar{rem:#1}
                      \label{rem:#1} \em }
\newcommand {\erem}   {\end{remark}}

\newcommand {\beq}[1]{\mymarginpar{eq:#1}
                      \begin{equation}
                      \label{eq:#1} }

\newcommand {\beqno}[1]{\mymarginpar{eq:#1}
                      \begin{eqnarray}
                      \nonumber}

\newcommand {\eeq}       {\end{equation}}
\newcommand {\eeqno}       { && \end{eqnarray}}

\newcommand {\bear}[1]{\mymarginpar{eq:#1}
                       \begin{eqnarray}
                       \label{eq:#1} }

\newcommand {\bearno}[1]{\mymarginpar{eq:#1}
                       \begin{eqnarray}
                       \nonumber}

\newcommand {\eear}{\end{eqnarray}}
\newcommand {\eearno}{\end{eqnarray}}
\newcommand {\bsel}{\left \{ \begin{array}{cl}}
\newcommand {\esel}{\end{array} \right.}

\newcommand {\bmat}[1]{\left [ \begin{array}{#1}}
\newcommand {\emat}{\end{array} \right ]}
\newcommand {\bsec}[2]{\mymarginpar{sec:#2}
                       \section{#1}
                       \label{sec:#2} }

\newcommand {\bsubsec}[2]{\mymarginpar{sec:#2}
                       \subsection{#1}
                       \label{sec:#2} }

\def\R{I\kern-0.30em R}
\def\N{I\kern-0.30em N}
\def\P{I\kern-0.30em P}

\begin{document}
\begin{CJK*}{UTF8}{bsmi}
\title{InterAct: Exploring the Potentials of ChatGPT as a Cooperative Agent}

\author{\IEEEauthorblockN{1\textsuperscript{st} Given Name Surname}
	\IEEEauthorblockA{\textit{dept. name of organization (of Aff.)} \\
		\textit{name of organization (of Aff.)}\\
		City, Country \\
		email address or ORCID}
	\and
	\IEEEauthorblockN{2\textsuperscript{nd} Given Name Surname}
	\IEEEauthorblockA{\textit{dept. name of organization (of Aff.)} \\
		\textit{name of organization (of Aff.)}\\
		City, Country \\
		email address or ORCID}
	\and
	\IEEEauthorblockN{3\textsuperscript{rd} Given Name Surname}
	\IEEEauthorblockA{\textit{dept. name of organization (of Aff.)} \\
		\textit{name of organization (of Aff.)}\\
		City, Country \\
		email address or ORCID}
	\and
	\IEEEauthorblockN{4\textsuperscript{th} Given Name Surname}
	\IEEEauthorblockA{\textit{dept. name of organization (of Aff.)} \\
		\textit{name of organization (of Aff.)}\\
		City, Country \\
		email address or ORCID}
	\and
	\IEEEauthorblockN{5\textsuperscript{th} Given Name Surname}
	\IEEEauthorblockA{\textit{dept. name of organization (of Aff.)} \\
		\textit{name of organization (of Aff.)}\\
		City, Country \\
		email address or ORCID}
	
}

\author{Po-Lin Chen and
		Cheng-Shang~Chang,~\IEEEmembership{Fellow,~IEEE}\\
		\IEEEcompsocitemizethanks{\IEEEcompsocthanksitem
		The authors are with the Institute of Communications Engineering, National Tsing Hua University, Hsinchu 300044, Taiwan R.O.C. Email:  b220954335@gmail.com; cschang@ee.nthu.edu.tw.
		\protect\\
	}
	\thanks{
 This work was supported in part by the National Science and Technology, Taiwan, under Grant 111-2221-E-007-045-MY3, and in part by Qualcomm Technologies under Grant SOW NAT-487844-2. }}

\maketitle

\begin{abstract}
This research paper delves into the integration of OpenAI's ChatGPT into embodied agent systems, evaluating its influence on interactive decision-making benchmark. Drawing a parallel to the concept of people assuming roles according to their unique strengths, we introduce InterAct. In this approach, we feed ChatGPT with varied prompts, assigning it a numerous roles like a checker and a sorter, then integrating them with the original language model. Our research shows a remarkable success rate of 98\% in AlfWorld, which consists of 6 different tasks in a simulated household environment, emphasizing the significance of proficient prompt engineering. The results highlight ChatGPT's competence in comprehending and performing intricate tasks effectively in real-world settings, thus paving the way for further advancements in task planning.

\end{abstract}


\section{Introduction}
\label{sec:introduction}

The advent of large language models (LLMs), underpinned by transformative advancements in natural language processing (NLP), has stimulated a revolution across a wide range of applications. Exemplified by models such as Transformer\cite{vaswani2017attention}, T5\cite{raffel2020exploring}, GPT-4\cite{openai2023gpt}, these language models have achieved impressive results in diverse tasks like paragraph summary, language translation, and code optimization. These achievements can be attributed to their ability to absorb and process massive amounts of data, making sense of the patterns and structures within the text.

ChatGPT\cite{chatGPT} is an AI language model created by OpenAI, which has been trained using a combination of pretraining and fine-tuning with human feedback. This advanced model is built on Transformer model, enabling it to produce responses that closely resemble human language. By undergoing extensive training on vast volumes of text data, ChatGPT excels in understanding and generating text in various languages and fields, answering queries, and engaging in dialogues. Unlike its predecessors that operate primarily based on a single prompt, ChatGPT combines text generation with code synthesis, thereby significantly enhancing its interactive abilities.

In this paper, we assess the ability of ChatGPT to make decisions within the context of an AlfWorld simulated environment\cite{shridhar2020AlfWorld}. The aim is to understand the model's proficiency in absorbing and processing data to make rational decisions. Scholarly works such as ReAct\cite{yao2022react} and Reflexion\cite{shinn2023reflexion} showcase the decision-making, action-initiation, and reflective powers of LLMs, paving the way for remarkable progress in a range of text-based performance metrics. However, they all utilize a single language model (InstructGPT) which, despite numerous iterations of thought and reflection, often repeatedly commits the same mistakes. In this research, we devise a novel model, InterAct, which is founded on the architecture of the ReAct model \cite{yao2022react}.  It undergoes alterations in prompt formulations, incorporates different ChatGPT for support. In particular, we add a {\em checker} module to tackle the issue of object misidentification. The initial basic prompt has also been revised to bolster InterAct's capabilities in constructing comprehensive search paths. This approach effectively addresses the previously mentioned shortcomings of the ReAct model. Consequently, this approach yielded a success rate of 98\% in this benchmark, a significant improvement from the base ReAct agent's accuracy of 75\%. These experiments provide critical insights into the potential benefits and limitations of implementing ChatGPT in AI-driven systems and technologies.

In conclusion, the main insight of the paper is the advancement of AI language models like ChatGPT presents an exciting opportunity to revolutionize and reshape our interaction with technology. By leveraging these models, we can build more intuitive, responsive, and smart technologies that can effectively understand and respond to human requirements. The key contributions of our research are summarized below:
\begin{description}

\item[(1)] We introduce InterAct, an improved method where each agent, like ChatGPT, can showcase unique abilities, adeptly rectifying the limitations found in the ReAct model, such as object misidentification and inefficient planning.

\item[(2)] We have designed new trajectory prompts that enable the agent to flawlessly locate items during its search process.

\item[(3)] In a decision-making test within the AlfWorld simulated environment, InterAct demonstrated a 98\% success rate, significantly higher than the 75\% accuracy of the base ReAct agent, suggesting its potential benefits in AI-centric systems and technologies.

\end{description}


\bsec{Related work}{Relatedwork}

\textbf{ Dominance of Transformers for Robots}\hspace{1em}   Transformers have emerged as the dominant architecture in various fields. Initially prominent in NLP\cite{fedus2022switch}, \cite{brown2020language}, \cite{zhuang2021robustly}, they have now extended their influence to include vision-based tasks \cite{liu2021swin}, \cite{liang2021swinir} and even reinforcement learning \cite{chen2021decision}, \cite{lee2022multi}.
In the realm of robotics, Transformers have found practical applications in diverse areas such as path planning \cite{alexis2015uniform}, \cite{chaplot2021differentiable}, object recognition \cite{he2022masked}, and grasping\cite{park2018classification}.

One notable example is RT-1\cite{brohan2022rt}, which takes the utilization of Transformers that takes images from a robot’s camera and natural language task instructions as inputs and directly outputs tokenized actions. RT-1 can also acquire new skills by observing other robots' experiences, opening opportunities for enhanced robot capabilities through multi-robot datasets. Another instance is SayCan\cite{ahn2022can}, a study conducted by Google's AI team and Everyday Robots. This research employs PaLM\cite{chowdhery2022palm} and an affordance function to empower robots to carry out complex tasks based on natural language instructions. The resulting system, PaLM-SayCan, transforms user instructions into actionable plans for the robot. Inner Monologue\cite{huang2022inner} has made further advancements by incorporating injected feedback from the environment. The work in \cite{huang2022language} demonstrated that even without any training, sizable language models can be effectively prompted to produce credible action plans driven by goals. They also suggested multiple techniques to enhance the model's ability to generate executable outputs, all without the need for invasive probing or modifications to the underlying model.

\textbf{GPT for Robotics}\hspace{1em} Moreover, recent publications, including \cite{vemprala2023ChatGPT}, \cite{wake2023chatgpt}, and \cite{lu2023agi}, have successfully incorporated models such as ChatGPT and GPT3.5 into the realm of robotics applications. These advancements facilitate interaction between the models and the environment or users, allowing for the correction of the robot's behavior. These papers showcase various prompts and outline a pipeline for the implementation of ChatGPT in robotics tasks. Additionally, they conduct experimental evaluations to assess ChatGPT's capability to execute a wide range of robotics tasks while striving to bridge the gap between natural language and actionable robot actions.

\textbf{LLM for Robotics reasoning}\hspace{1em} The process of reasoning in robotics involves breaking down complex tasks into simpler subtasks that can be more easily solved by the LLM itself or with the aid of tools. Various approaches \cite{chung2022scaling}, \cite{wei2022emergent} have been introduced to enable natural language agents to select their next action in text-based environments.

One prominent approach is Chain-of-thought (CoT) reasoning, as proposed in \cite{wei2022chain}. This approach leverages emergent properties, such as reasoning and commonsense, to solve tasks through multiple steps. It enables the LLM to reason through a series of intermediate actions, leading to the desired outcome.

Another approach called faithful reasoning, introduced in \cite{creswell2022faithful}, decomposes multi-step reasoning into three distinct steps, each handled by a dedicated LLM. By dividing the task into these steps, faithful reasoning facilitates the LLM's ability to tackle complex computations effectively. Similar approaches like Scratchpad \cite{nye2021show}, which involves fine-tuning an LLM on intermediate computation steps, resulting in improved performance on multi-step computation problems.

The Describe, Explain, Plan, and Select (DEPS) approach, introduced in \cite{wang2023describe}, specifically developed to tackle the unique challenges of planning in open-ended environments such as Minecraft. This innovative system adeptly manages intricate tasks that demand meticulous, multi-step reasoning, effectively prioritizing sub-goals according to the agent's proximity. Notably, DEPS has exhibited remarkable results in enhancing the success rate of Minecraft tasks by offering insightful explanations for errors encountered during sub-task execution. As a groundbreaking planning agent, DEPS has achieved an unprecedented positive success rate in conquering the formidable ObtainDiamond task, marking a significant milestone in the field.

A different strategy called DERA\cite{nair2023dera} presents an alternative approach by structuring a dialogue as a conversation between two agent types: "Researcher" and "Decider." The Researcher agent analyzes information and identifies key components of the problem, while the Decider agent autonomously combines the Researcher's insights and makes judgments on the final output. This approach has demonstrated notable enhancements compared to the baseline performance of GPT-4\cite{openai2023gpt} in evaluations conducted by human experts and quantitative metrics. Particularly, DERA has showcased significant advancements in safety-critical domains like healthcare.

Additionally, the studies by \cite{shinn2023reflexion}, \cite{madaan2023self} have also incorporated reflection actions into the model. These reflection actions allow the model to refine its actions based on feedback received during the execution of tasks. By iteratively adjusting its actions and incorporating self-feedback, the model can improve its decision-making process and adapt to changing conditions.

Our research aims to provide additional evidence supporting the effectiveness of ChatGPT in language-conditioned robotic learning simultaneously introducing novel architectures that facilitate reasoning through the coordination of various roles performed by LLMs.

\bsec{Method: InterAct Structure}{construction}

In this section, we use the AlfWorld benchmark to test ChatGPT's reasoning capabilities, examining how it accomplishes household tasks step by step when provided only with a few-shot example. We will use not only ChatGPT but also a similar language model called InstructGPT (text-davinci-002). InstructGPT is particularly adept at tasks demanding succinct responses or benefiting from k-shot examples. In this particular task, unlike the previous demostration, the model is required to integrate task-oriented actions with verbal reasoning. The model needs to possess the ability to think and reason like a human. When faced with dead ends , the model should be capable of adjusting its planning based on logical reasoning.

\bsubsec{AlfWorld Dataset}{dataset}

AlfWorld is a suite of text-based environments that challenge an agent to solve multi-step tasks in a variety of interactive environments with ALFRED\cite{shridhar2020alfred} benchmark. The ALFRED benchmark focuses on tasks that require an agent to accomplish high-level goals in a simulated household environment by navigating and interacting through text-based actions. In AlfWorld, there are six types of tasks that challenge the agent's ability to plan, track subgoals, and explore systematically.

For example, a task in AlfWorld could be to "examine a paper under a desklamp." To achieve this goal, the agent needs to navigate to specific locations within the simulated household and interact with objects using text commands. The agent might need to issue commands like "go to coffeetable 1," "take paper 2," and "use desklamp 1" to complete the task.

The complexity of the tasks in AlfWorld is intentionally designed to be challenging. Task instances can have more than 50 locations and may require an expert policy more than 50 steps to solve. This complexity encourages the agent to effectively plan its actions, keep track of subgoals, and explore the environment systematically. For example, the agent may need to check all desks one by one to find the desklamp.

One of the challenges presented in AlfWorld is the need to determine likely locations for common household items. For instance, a desklamp is likely to be found on desks, shelves, or dressers. This aspect of the environment provides an opportunity for language models like LLMs to leverage their pretrained commonsense knowledge to make informed decisions about the likely locations of objects.

In each environment of AlfWorld, the agent has the option to select an action from a list of permissible actions, denoted as $A_t$ at time step t. Upon executing an action, the agent receives an observation, $O_t$, and a reward, $R(s_t, a_t)$, from the environment, which then determines the next state of the agent.

AlfWorld offers a diverse set of six tasks and a total of over 3000 unique environments. These environments test the agent's ability to understand the task at hand, formulate a sequential plan consisting of subtasks, and carry out the necessary actions within the given environment. In our trials, we utilize the ReAct problem-solving strategy\cite{yao2022react}, which has demonstrated superior performance across a wide array of sequential decision-making tasks. ReAct is a strategy that allows the agent to reason and act by articulating its current thoughts and performing actions based on these thoughts. At each time step, the agent has the option to execute $<think>$: thought action to verbalize its internal thought process, or $<action>$: to induce a response from the environment. The set of possible actions in each state is not explicitly defined, providing the agent with full autonomy in determining its next moves. To prevent syntactic errors, we provide the agent with two domain-specific few-shot trajectories.

\bsubsec{Model architecture}{architecture}

\begin{figure}[htbp]
\centering
\includegraphics[width=1\columnwidth]{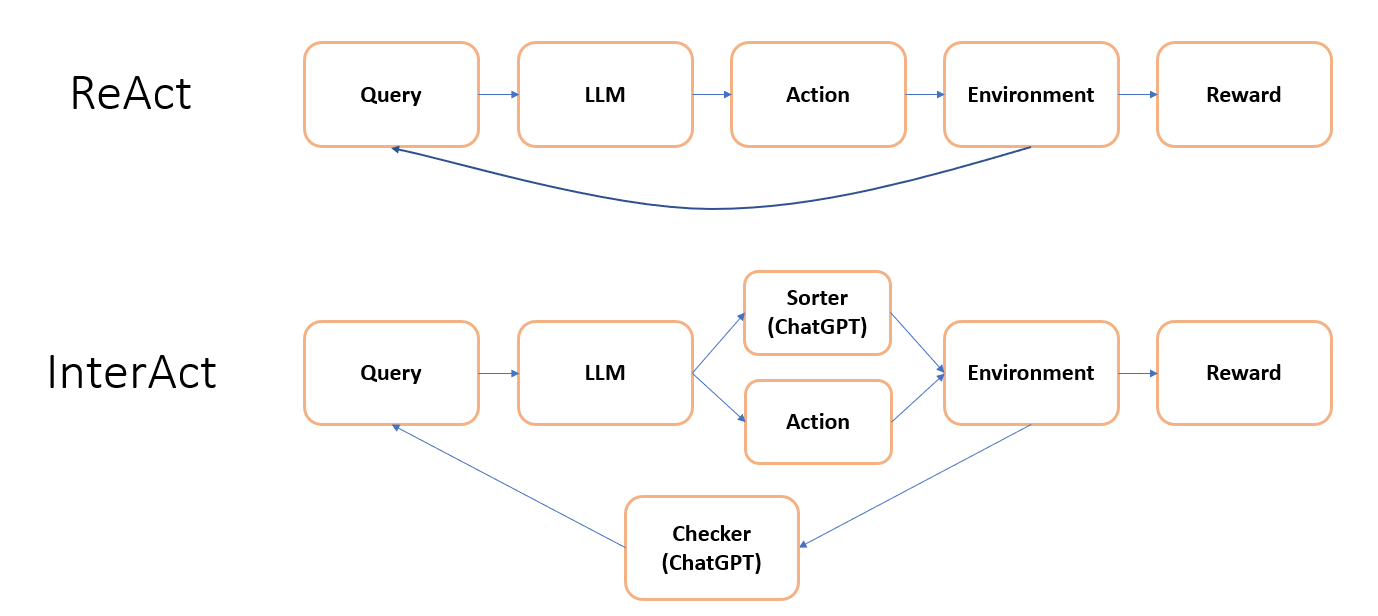}
\caption{The architecture of both ReAct and InterAct. InterAct involves the integration of LLM with various agents to facilitate smoother interaction with the environment.}	
\label{fig:model architecture}
\end{figure}

We introduced a novel model called InterAct, which is built upon the foundation of ReAct. The architectural diagram of InterAct can be observed in \rfig{model architecture}. While ReAct has demonstrated impressive accuracy in diverse decision-making and knowledge-intensive tasks, it occasionally encounters common errors, including Perception Error, Object Misidentification, and Inefficient Planning. In simpler terms, although ReAct achieves state-of-the-art performance overall, there exists a small subset of tasks that remain unsolved due to minor imperfections in a single model.

To address these challenges, InterAct leverages the combined strength of agents with distinct purposes, such as checker and sorter, to enhance the areas where ReAct is susceptible to errors. In addition, we have modified the original basic prompt to enhance InterAct's ability to plan comprehensive search paths when looking for multiple items, ensuring that no possible locations are overlooked. This optimization greatly improves the efficiency of the tasks being performed.

\begin{figure}[htbp]
\centering
\includegraphics[width=1\columnwidth]{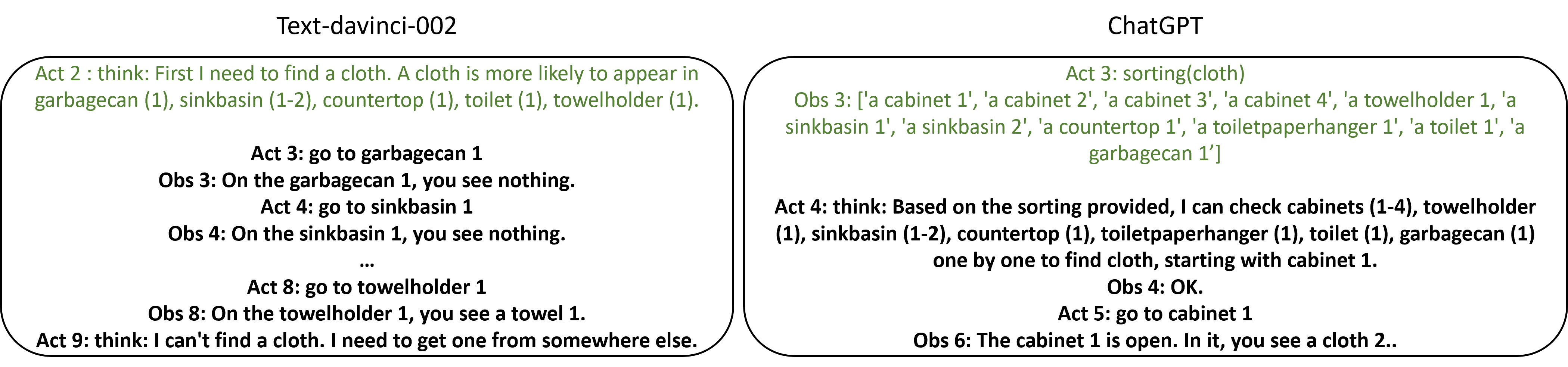}
\caption{The left image was generated using text-davinci-002 for search ranking, while the right image was generated using ChatGPT. It can be observed that ChatGPT exhibits higher logical reasoning in finding objects compared to text-davinci-002.}	
\label{fig:Sorter}
\end{figure}

\textbf{Sorter}\hspace{1em}  When processing environmental data, ReAct initially needs to determine the likelihood of objects appearing in specific locations. However, this ranking process often falls short, leading to less efficient planning. This inefficiency may arise from the fact that the  the InstructGPT model (text-davinci-002) is not sufficiently trained in factual knowledge and common-sense reasoning. On the other hand, ChatGPT has been fine-tuned using Reinforcement Learning with Human Feedback (RLHF) and has demonstrated a more nuanced understanding of various situations. It excels at making well-informed decisions, as depicted in \rfig{Sorter}. To improve the efficiency of predicting object locations, we integrate ChatGPT as a decision-making component. Whenever ReAct requires this procedure, it can autonomously utilize ChatGPT, thus enhancing the effectiveness of its object search operations.

\begin{figure}[htbp]
\centering
\includegraphics[width=0.9\columnwidth]{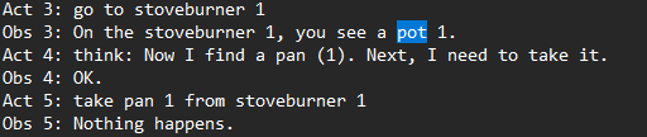}
\caption{Object Misidentification. In this scenario, the objective is to locate a pan; however, ReAct mistakenly misidentifies another object as the pan.}	
\label{fig:object misidentification}
\end{figure}

\textbf{Checker}\hspace{1em}  Another issue with text-davinci-002 is that it tends to mistakenly categorize similar objects as the same. For example, it might treat a pan and a pot as identical items, leading to the problem of Object Misidentification, as depicted in \rfig{object misidentification}. To address this issue, we employ ChatGPT as a checker by providing it with appropriate prompts. We have observed that ChatGPT can successfully distinguish between similar objects. Furthermore, we utilize the results from this checker as observations and feed them back to the LLM, as illustrated in \rfig{model architecture}. This approach helps us resolve the problem related to object misidentification.

\begin{figure}[htbp]
\centering
\includegraphics[width=1\columnwidth]{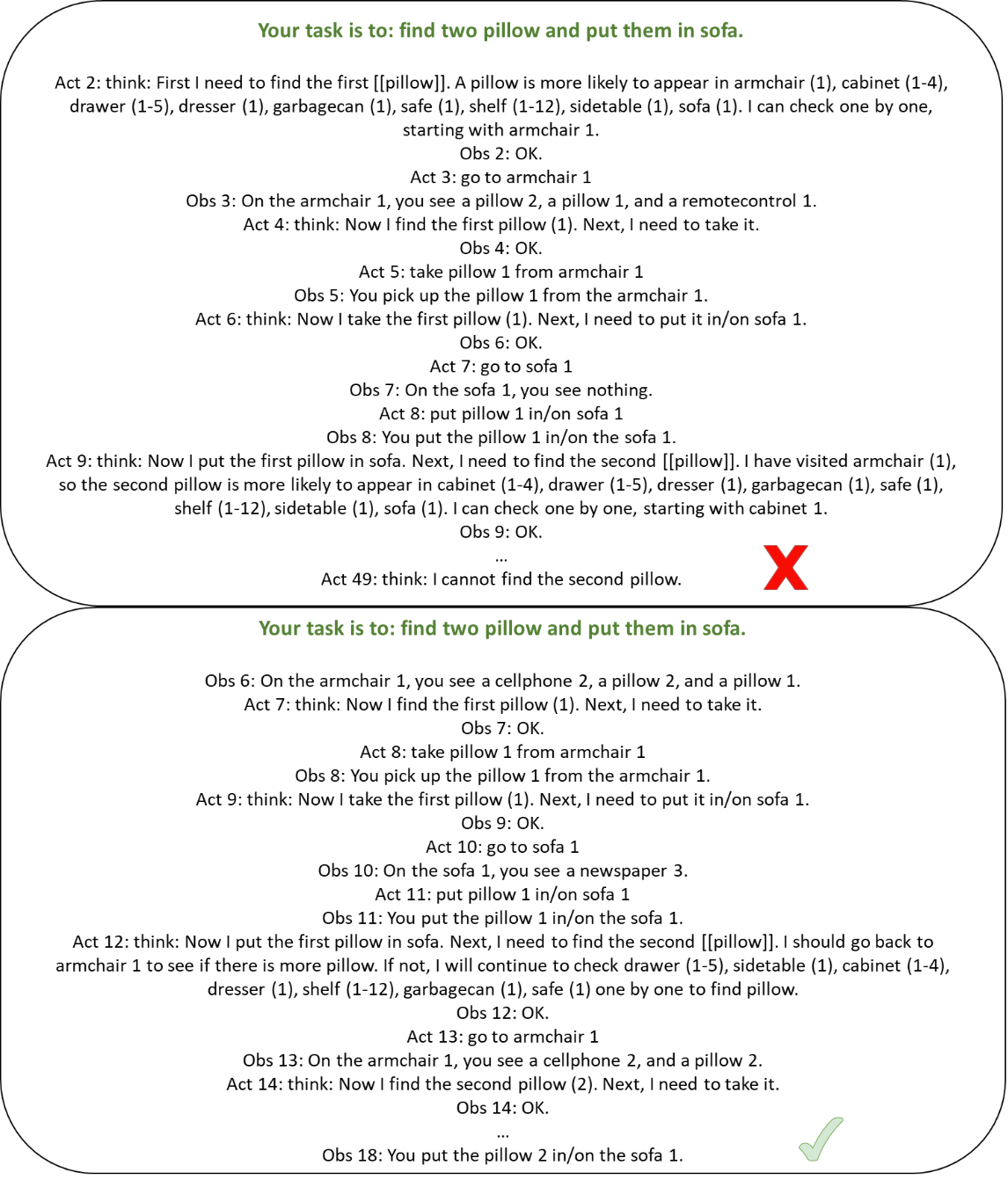}
\caption{Trajectory planning. In the initial scenario, the agent fails to retrieve the second pillow from the armchair after placing the first pillow on the sofa. Consequently, the agent cannot find the second pillow, resulting in an incomplete task. In the revised scenario, InterAct addresses this issue by considering the future search trajectory. It prioritizes returning to the armchair to search for the second pillow before exploring the other areas. This approach improves the chances of successfully locating the second pillow and completing the task.}	
\label{fig:new trajectory}
\end{figure}

\textbf{Trajectory planning}\hspace{1em}  In the AlfWorld environment, we encountered a
mission type named ”pick 2”, where the agent is required to find two identical objects. We observed that ReAct alone tends to forget its previous locations, resulting in inefficient trajectories characterized by frequent revisits to the same place. In some instances, this led to hallucinations, defined as consecutive identical actions with the environment responding similarly. To address this issue, we made changes to the original model's prompt. After finding the first object and placing it in the corresponding receptacle, we allow the model to autonomously generate a trajectory while ensuring that this path does not overlook areas where the second object might be present, as shown in \rfig{new trajectory}.

More details about prompts, we refer the reader to Appendix A.

\bsec{Evaluation}{evaluation}

In this section, we present a comparative analysis of the performance enhancement provided by the helpers (sorter or checker) and the new trajectory planning when compared to the baseline model. Our findings demonstrate that InterAct consistently outperforms ReAct on AlfWorld (as shown in Table \ref{tb: interact}) across all tasks. On AlfWorld, the top-performing InterAct trial achieves an impressive average success rate of 98\%, falling short in only 2 out of 134 tasks.  This performance is significantly better than the best trials of ReAct (73\%) and BUTLER (37\%). Indeed, InterAct has demonstrated exceptional proficiency in handling these tasks, as evidenced by achieving a 100\% success rate in four out of the six tasks. This performance showcases InterAct's remarkable ability to effectively manage and succeed in various tasks. Notably, even when ReAct is augmented only with a checker or sorter, the overall average performance surpasses that of ReAct without helpers by a significant margin. The tasks that show the most substantial improvement are "pick2" and "clean," with an approximate gain of 47\% and 41\%. From a qualitative standpoint, we observed that ReAct, without any helper, faces difficulties in accurately determining the presence of items in a specific location or employing ineffective search strategies.

\begin{table}[!htbp]
    \begin{center}
\renewcommand\arraystretch{1.5}
    \caption{AlfWorld task-specific success rates ($\%$).}
	\scalebox{1}{
\begin{tabular}{c|c|c|c|c|c|c|c}
\hline
\hline
{ Method }&  Pick & Clean & Heat & Cool & Look & Pick2 & { All }\\
\hline
{BUTLERg} & 33 & 6  & 70 & 76 & 17 & 12  & {46} \\
{BUTLER}  &  65 & 39 & 83 & 76 & 55 & 24  & {57}\\
\hline
{Act} &  88 & 41 & 76 & 67 & 73 & 43  & {46} \\
{ReAct}  &88 & 55  & 90&  81 & 75 & 53  & {73}\\
\hline
{ReAct+checker} &  85 & 81 & 100 & 87 & 92 & 75  & {86} \\
{ReAct+sorter}  &  84 & 76 & 88 & 73 & 80 & 67  & {78}\\
{InterAct} &  \textbf{100} & \textbf{96} & \textbf{100} & \textbf{94} & \textbf{100} & \textbf{100}   & {\textbf{98}} \\

\hline
\hline
    \end{tabular}\label{tb: interact}
}
    \end{center}
\end{table}

\bsec{Discussion and Limitations}{correlatedLDPC}

\bsubsec{Scalability of InterAct}{discussion1}

Our InterAct model is scalable and adaptable to different datasets and scenarios. For instance, if there's a need for a feature similar to 'memories,' we can develop an interpreter to describe the current path, among other things, without having to train numerous different language models. This is possible because ChatGPT serves as an excellent backbone for such extensions.

\bsubsec{Error assessment with a supervisor module}{discussion3}

Despite achieving an impressive average performance of 98\% on the AlfWorld dataset, our analysis of failed trajectories uncovered certain limitations. One notable drawback is the model's heavy reliance on prompt completeness within InterAct. When our examples contain missing or unaddressed components, the model fails to detect these errors, resulting in repetitive actions, even for trivial mistakes. To overcome this issue, we explored the possibility of using an alternative ChatGPT model as a supervisor to identify such errors. However, it's important to acknowledge that the accuracy of the supervisor's judgment cannot be guaranteed, and there may be occasional misidentifications leading to "action errors."

In order to tackle the challenge of error detection, we conducted a comparison between ChatGPT and GPT-4. The results demonstrated a significant improvement in error detection performance with GPT-4. Unfortunately, GPT-4 is currently unavailable as an open-source model and cannot be accessed free of charge. Conducting extensive simulations using GPT-4 requires funding support.

\bsubsec{Insufficiency of the dataset}{discussion4}

While AlfWorld is a valuable platform for assessing AI performance, it has certain limitations. Primarily, it encompasses only six types of tasks, and even within these categories, the task quantity is quite limited. These restrictions neither fully test nor make optimal use of the AI systems' capabilities. If we move to an environment offering a larger range and diversity of tasks, as well as a broader and more varied set of locations, our model will still need improvement to maintain its current level of accuracy. This aspect will be our focus for future research.

\bsec{Conclusion}{numerical}

Our research is centered on enhancing the task planning capabilities of large language models. We developed a new model, InterAct, built upon the framework of the ReAct model. InterAct is a culmination of various 'helpers' (like checkers and sorters) and aims to improve upon the existing trajectory. We evaluated this framework in the AlfWorld simulated environment, where it showed a substantial increase in decision-making accuracy, soaring from 75\% to an impressive 98\%. This highlights the vast potential of these models in AI-driven systems and technologies.

In essence, this study underscores the revolutionary potential of AI language models like ChatGPT and their pivotal role in shaping future real-world interactions. As we continue to delve into their capabilities, we are on the cusp of a new technological era marked by not only intelligence but also intuitiveness and responsiveness to human needs.

\appendices

\section{AlfWorld Experiment details}


\newtcolorbox{mybox}[2][]
  {colback = gray!10!white, colframe = gray!75!black, fonttitle = \bfseries,
    colbacktitle = gray!90!black, enhanced,
    attach boxed title to top center={yshift=-2mm},
    title=#2,#1,fontupper=\footnotesize	\fontfamily{ppl}\selectfont,breakable}

Here's an example in the prompt of InterAct in heat task. In the prompt, we enclose the items that need to be found within double square brackets, so that they can be fed to the checker and sorter later. InterAct will search for the items based on the sorting result. Each time we arrive at a location, we ask the checker to find the required item. If it is found, we consider the things present at that location as observations and return them to the LLM. If the item is not found, we ask the checker to return the message "Object is not here.".

\begin{mybox}{$<$heat$>$ Prompt}
\textbf{You are in the middle of a room. Looking quickly around you, you see a cabinet 10, a cabinet 9, a cabinet 8, a cabinet 7, a cabinet 6, a cabinet 5, a cabinet 4, a cabinet 3, a cabinet 2, a cabinet 1, a coffeemachine 1, a countertop 3, a countertop 2, a countertop 1, a diningtable 1, a drawer 6, a drawer 5, a drawer 4, a drawer 3, a drawer 2, a drawer 1, a fridge 1, a garbagecan 1, a microwave 1, a sinkbasin 1, a stoveburner 4, a stoveburner 3, a stoveburner 2, a stoveburner 1, and a toaster 1.}

\textbf{Your task is to: heat some egg and put it in diningtable.}

$>$ think: To solve the task, I need to find and take an egg, then heat it with microwave, then put it in diningtable.

OK.

$>$ think: First I need to find an [[egg]].

OK.

$>$ sorting(egg)

[a fridge 1, a countertop 1, a countertop 2, a countertop 3, a diningtable 1, a sinkbasin 1, a garbagecan 1, a microwave 1, a cabinet 1, a cabinet 2, a cabinet 3, a cabinet 4, a cabinet 5, a cabinet 6, a cabinet 7, a cabinet 8, a cabinet 9, a cabinet 10, a stoveburner 1, a stoveburner 2, a stoveburner 3, a stoveburner 4, a drawer 1, a drawer 2, a drawer 3, a drawer 4, a drawer 5, a drawer 6, a coffeemachine 1, a toaster 1]

$>$ think: Based on the sorting provided, I can check fridge (1), countertop (1-3), diningtable (1), sinkbasin (1), garbagecan (1), microwave (1), cabinet (1-10),  stoveburner (1-4), drawer (1-6), coffeemachine (1), toaster (1) one by one to find egg, starting with fridge 1.

OK.

$>$ go to fridge 1

The fridge 1 is closed.

$>$ open fridge 1

egg is not here.

$>$ go to countertop 1

egg is not here.

$>$ go to countertop 2

egg is not here.

$>$ go to countertop 3

On the countertop 3, you see a bowl 1, a butterknife 1, a egg 2, a kettle 2, a plate 1, a sink 1, and a spatula 2.

$>$ think: Now I find an egg (2). Next, I need to take it.

OK.

$>$ take egg 2 from countertop 3

You pick up the egg 2 from the countertop 3.

$>$ think: Now I take an egg (2). Next, I need go to a microwave (1) and heat it.

OK.

$>$ go to microwave 1

The microwave 1 is closed.

$>$ heat egg 2 with microwave 1

You heat the egg 2 using the microwave 1.

$>$ think: Now I heat an egg (2). Next, I need to put it in/on diningtable 1.

OK.

$>$ go to diningtable 1

On the diningtable 1, you see a apple 2, a bread 3, a egg 1, a kettle 1, a knife 1, a mug 1, a papertowelroll 1, a peppershaker 2, a potato 1, a soapbottle 1, and a spatula 1.

$>$ put egg 2 in/on diningtable 1

You put the egg 2 in/on the diningtable 1.

\end{mybox}

In the "pick2" task, we have incorporated new trajectory planning. After locating and placing the first object in the appropriate receptacle, we enable the model to independently generate a trajectory, ensuring that the path does not overlook areas where the second object may be located.

\begin{mybox}{$<$picktwo$>$ Prompt}
\textbf{You are in the middle of a room. Looking quickly around you, you see a armchair 2, a armchair 1, a bed 1, a countertop 1, a diningtable 1, a drawer 2, a drawer 1, a dresser 1, a laundryhamper 1, and a sidetable 1.}

\textbf{Your task is to: put two creditcard in dresser.}

$>$  think: To solve the task, I need to find and take the first creditcard, then put it in dresser, then find and take the second creditcard, then put it in dresser.

OK.

$>$  think: First I need to find the first [[creditcard]].

OK.

$>$  sorting(creditcard)

[countertop 1, diningtable 1, sidetable 1, armchair 1, armchair 2, bed 1, drawer 1, drawer 2, dresser 1, laundryhamper 1]

$>$  think: Based on the sorting provided, I can check coutertop (1), diningtable (1), sidetable (1),  armchair (1-2), bed (1), drawer (1-2), dresser (1), laundryhamper (1) one by one to find creditcard, starting with countertop 1.

OK.

$>$  go to countertop 1

creditcard is not here.

$>$  go to diningtable 1

On the diningtable 1, you see a cellphone 2, a creditcard 4, a creditcard 3, a creditcard 2, a mirror 1, a pencil 2, and a pencil 1.

$>$  think: Now I find the first creditcard (2). Next, I need to take it.

OK.

$>$  take creditcard 2 from diningtable 1

You pick up the creditcard 2 from the diningtable 1.

$>$  think: Now I take the first creditcard (2). Next, I need to put it in/on dresser 1.

OK.

$>$  go to dresser 1

On the dresser 1, you see a mug 1, and a television 1.

$>$  put creditcard 2 in/on dresser 1

You put the creditcard 2 in/on the dresser 1.

$>$  think: Now I put the first creditcard in dresser. Next, I need to find the second [[creditcard]]. I should go back to diningtable 1 to see if there is more creditcard. If not, I will continue to check sidetable (1),  armchair (1-2), bed (1), drawer (1-2), dresser (1), laundryhamper (1) one by one to find creditcard.

OK.

$>$  go to diningtable 1

On the diningtable 1, you see a cellphone 2, a creditcard 4, a creditcard 3, a mirror 1, a pencil 2, and a pencil 1.

$>$  think: Now I find the second creditcard (3). Next, I need to take it.

OK.

$>$  take creditcard 3 from diningtable 1

You pick up the creditcard 3 from the diningtable 1.

$>$  think: Now I take the second creditcard (3). Next, I need to put it in/on dresser 1.

OK.

$>$  go to dresser 1

On the dresser 1, you see a creditcard 2, a mug 1, and a television 1.

$>$  put creditcard 3 in/on dresser 1

You put the creditcard 3 in/on the dresser 1.

\end{mybox}
\end{CJK*}
\end{document}